# Efficient Autonomous Navigation for Terrestrial Insect-Machine Hybrid Systems

H. Duoc Nguyen [a], V. Than Dung [a], Hirotaka Sato [a, *], and T. Thang Vo-Doan [b, **]

[a] *School of Mechanical & Aerospace Engineering, Nanyang Technological University, 50 Nanyang Avenue, 639798, Singapore*

[b] *Institute of Biology I, University of Freiburg; Hauptstrasse 1, Freiburg, 79104, Germany*

[*, **] Corresponding authors. *Email addresses*: hirosato@ntu.edu.sg (Hirotaka Sato), vodoan@bio.uni-freiburg.de (T. Thang Vo-Doan)

**Abstract**

While bio-inspired and biomimetic systems draw inspiration from living materials, biohybrid systems incorporate them with synthetic devices, allowing the exploitation of both organic and artificial advantages inside a single entity. In the challenging development of centimeter-scaled mobile robots serving unstructured territory navigations, biohybrid systems appear as a potential solution in the forms of terrestrial insect-machine hybrid systems, which are the fusion of living ambulatory insects and miniature electronic devices. Although their maneuver can be deliberately controlled via artificial electrical stimulation, these hybrid systems still inherit the insects' outstanding locomotory skills, orchestrated by a sophisticated central nervous system and various sensory organs, favoring their maneuvers in complex terrains. However, efficient autonomous navigation of these hybrid systems is challenging. The struggle to optimize the stimulation parameters for individual insects limits the reliability and accuracy of navigation control. This study overcomes this problem by implementing a feedback control system with an insight view of tunable navigation control for an insect-machine hybrid system based on a living darkling beetle. Via a thrust controller for acceleration and a proportional controller for turning, the system regulates the stimulation parameters based on the instantaneous status of the hybrid robot. While the system can provide an overall success rate of ~71% for path-following navigations, fine-tuning its control parameters could further improve the outcome's reliability and precision to up to ~94% success rate and ~1/2 body length accuracy, respectively. Such tunable performance of the feedback control system provides flexibility to navigation applications of insect-machine hybrid systems.

## 1. Introduction

Terrestrial insect-scale mobile robots have become prominent candidates for post-disaster search-and-rescue missions. Their tiny size and light weight would help them easily penetrate deep into the rubbles of collapsed buildings without causing additional collapses. While there are growing efforts to achieve insect-level autonomy in these robots, it is still a challenge to match their natural-born counterparts, i.e., living ambulatory insects. While control autonomy was achieved in various insect-scale mobile robots (Chen et al. 2020; de Rivaz et al. 2018; Goldberg et al. 2018; St. Pierre and Bergbreiter 2019; Yang et al. 2020), power autonomy was demonstrated only in a few platforms, like HAMR-F (Goldberg et al. 2018) or Robeetle (Yang et al. 2020). Furthermore, although inverted and vertical climbing was demonstrated (Chen et al. 2020; de Rivaz et al. 2018), maneuvering across complex terrains is still a conundrum for these artificial robots. However, the answers to these challenges might be found in the emerging ideas of biohybrid systems, which utilize the advantages of organic components in robotics applications (Anderson et al. 2020; Lee et al. 2022; Nitta et al. 2021; Webster-Wood et al. 2017; Yalikun et al. 2019). While bio-inspired and biomimetic systems seek insights from living materials, biohybrid systems incorporate them with synthetic devices (Anderson et al. 2020; Webster-Wood et al. 2017). This approach allows the hybrid systems to exploit both organic and artificial benefits from their two facets (Anderson et al. 2020; Webster-Wood et al. 2017). These systems were widely demonstrated in the forms of millimeter-scale actuators or aquatic/airborne miniature robots (Anderson et al. 2020; Lee et al. 2022; Nitta et al. 2021; Yalikun et al. 2019). In the case of insect-scale mobile robots, the biohybrid systems might be built by integrating living terrestrial insects (as robotics platforms) with synthetic electronic devices (as central control units). This integration would allow the inheritance of the insects' autonomy, including their perfect locomotory capabilities to crossover complex terrains, regulated by a top-notch inner control system (i.e., their central nervous system, CNS) (Bläsing and Cruse 2004; Ritzmann and Büschges 2007). Besides, the insects' fascinating biological body structure, which is made of natural soft actuators (muscles), durable exoskeleton, complex joints, integrated proprioceptors, and environmental sensors, will provide these living insect-based systems with both flexibility and robustness (Webster-Wood et al. 2017).

Such biohybrid systems are also known as insect-machine hybrid systems, cyborg insects, or insect biobots. These insect-machine hybrid systems are the fusion of a living insect platform and a miniature electronic device (or a backpack) employed to regulate the insect's locomotion via electrical stimulation (Li and Sato 2018; Maharbiz and Sato 2010). These hybrid systems will possess the potential to outperform the artificial insect-scale mobile robots in insect-like autonomy. More specifically, with a simple add-on electronic circuit, these systems can retain the outstanding locomotory skills of the insect platform to favor the complex and unpredictable post-calamity terrains (Bozkurt et al. 2016). At the same time, they can still provide various controllability with low power consumption (i.e., a few 100 µW) (Cao et al. 2014; Nguyen et al. 2020; Vo Doan et al. 2017). Moreover, insect-machine hybrid systems are environmentally friendly, have a low production cost, and allow a sustainable supply since most insect bodies are biodegradable and the insects are naturally reproducible.

Locomotion control of insect-machine hybrid systems can be realized via the electrical stimulation of their (peripheral or central) nervous or neuromuscular systems (Holzer and Shimoyama 1997; Latif and Bozkurt 2012; Sanchez et al. 2015; Sato et al. 2009; Sato et al. 2015). This stimulation-based control brings simplicity by freeing the electronic backpack from sophisticated tasks conventionally required for legged robots (e.g., coordinating multiple legs and joints (Luneckas et al. 2020)). The computational resource thus is saved for other essential tasks (e.g., victim detection (Tran-Ngoc et al. 2021) or localization (Cole et al. 2020)). Besides, the well-established stimulation methods, which are greatly diverse in both regulated movements and insect species, provide flexible choices to cope with the desired locomotion control from the diverse insect kingdom. For example, a variety of motions (e.g., directional turns, forward/backward/sideways walks) in darkling beetles (*Zophobas morio*) and cockroaches (*Periplaneta americana*, *Gromphadorhina portentosa*) can be induced by stimulating their sensory systems like antennae, elytra, and cerci (Dirafzoon et al. 2017; Latif and Bozkurt 2012; Nguyen et al. 2020; Tran-Ngoc et al. 2021; Vo Doan et al. 2017). These induced motions were discussed as reassembling the insects' behavioral responses when the same sensory organs were naturally evoked (Camhi and Tom 1978; Ye et al. 2003). Electrical stimulation of muscular systems can also be employed. For instance, stimulating a giant beetle's leg and flight muscles (*Mecynorrhina torquata*) reportedly regulated both its terrestrial and inflight motions (Cao et al. 2016; Vo-Doan et al. 2022). Similarly, the stimulation of neural sites also involves maneuvering insect-machine hybrid systems. For example, the flight initiation and cessation of giant beetles and the turning motion of cockroaches (*Blaberus discoidalis*) were obtained by stimulating their optic lobes (Sato et al. 2009) and prothoracic ganglion (Sanchez et al. 2015), respectively.

In the recent decade, the development of locomotion control gradually enabled the navigation control of insect-machine hybrid systems, especially in terrestrial species. The pioneer demonstrations were manual navigations, directing the insects along predetermined paths (Latif and Bozkurt 2012; Moore et al. 1998). Automatic navigation systems were then presented to evaluate the insects' response to their stimulation (Whitmire et al. 2013) or exhibited functionalities of the backpack (Whitmire et al. 2014). However, neither the systems' performances nor their control algorithms were introduced and analyzed in detail. Recently, a control system for autonomous navigation of terrestrial insect-machine hybrid systems exploring unknown and obstructed environments was demonstrated (Tran-Ngoc et al. 2021). Together with its pioneers, this system gradually brings insect-machine hybrid systems closer to their practical applications in search-and-rescue missions.

Despite such a progressive development, almost all contemporary navigations of terrestrial insect-machine hybrid systems encountered an identical issue, the high variation in the effect of electrical stimulation across individual insects and species. Several studies employed a fixed stimulus, commonly interpreted as an effective one, across different insect individuals. In addition, such an effective stimulus was altered between studies (Erickson et al. 2015; Latif and Bozkurt 2012; Whitmire et al. 2013). However, as each insect is physiologically unique, one's response to an identical stimulus will be unavoidably different from others. Thus, the fixity of electrical stimulation might cause a great deviation in the insects' response, consequently losing their controllability and worsening the navigation outcomes. This causality might explain the struggle of previous studies, in which the rate of successfully directing the insect-machine hybrid systems along the assigned paths was ~10-30% (Latif and Bozkurt 2012; Whitmire et al. 2013). Although these numbers did not represent the studies' significance (as navigation was not the focus), they undoubtedly imply an urge to resolve the issue of electrical stimulation. While adjusting the stimulus individually for each insect is possible, this procedure is tedious. The search for an optimal stimulus to produce a convergent locomotory response across different individuals was attempted (Erickson et al. 2015). Despite its moderate outcome (i.e., ~50% of the experimented insects reacted similarly), this approach was limited to one species, *Gromphadorhina portentosa*, and still untested in navigation control (Erickson et al. 2015).

In tandem with these insect-centric approaches, researchers are also looking for solutions from engineering perspectives, for example, the control algorithm for insect-machine hybrid systems. Many insect species reportedly exhibited graded locomotion responses driven by electrical stimulation (Cao et al. 2014; Latif et al. 2016; Nguyen et al. 2020; Vo Doan et al. 2017). Such graded reactions suggest a potential solution to overcome the demand for an optimal stimulus by utilizing a feedback control system in which the controllers can fine-tune the insects' desired responses (Cao et al. 2014; Li et al. 2018; Liu et al. 2022). For example, the elicited angular speed of darkling beetles (via their antennae stimulation) and the provoked thrust force of giant beetles (via the activation of their subalar muscles) were reportedly in direct and inversely proportional relationships, respectively, with the frequency of the electrical stimulation (Doan et al. 2015; Vo Doan et al. 2017). These relationships could be implemented to build a controller for regulating the insects' induced locomotory reactions and thus close the control loop for their navigation. Consequently, such a feedback control-based process would potentially resolve the issue of greatly deviated responses without knowing each insect's exact optimal parameters. The applicability of this approach was successfully demonstrated in the regulation of legs' motions (Cao et al. 2014) and straight-line flight control of giant flower beetles (*Mecynorrhina torquata*) (Li et al. 2018) or the accurate orientation adjustment of migratory locusts (*Locusta migratoria manilensis*) (Liu et al. 2022). However, neither the control system's precision nor its efficiency was reflected in the outcome of these works.

Thus, it is necessary to gain more in-depth insight into the role and performance of applying feedback control-based approaches for insect-machine hybrid systems, specifically for their automatic terrestrial navigations, as a pathway to actualize their practical applications.

Herein, this study demonstrates an ambulatory insect-machine hybrid system's feedback control-based navigation by implementing simple proportional controllers for its turning and forward motions to achieve its arbitrary path-following maneuver. The insect-machine hybrid system (Fig. 1) was assembled from a living darkling beetle (*Zophobas morio*, ~0.6 g) and a miniature wireless backpack (~0.25 g). A feedback control system was developed to automatically navigate the beetle along a predetermined sine curve (Fig. 2). The cores of this system included a proportional controller to steer the beetle precisely and a thrust controller to accelerate it forward rapidly. The former was derived from the beetle's graded response to its antennae stimulation, whereas the latter was inspired by its reaction to the elytra stimulation (Nguyen et al. 2020; Vo Doan et al. 2017). The performance of the designed feedback system was investigated by adjusting its control parameters. The adjustment allowed the system's navigation to attain a high success rate of up to 94% and a low path-tracking error of less than half of the beetle's body length, suggesting the implemented engineering control factors engulfed the biological differences between individual beetles. Also, this feedback control-based navigation was tunable by varying the control parameters, thus providing flexibility to its applications. As a result, the study displays the attainment of highly successful and precise automatic navigations for (terrestrial) insect-machine hybrid systems, thus verifying the significant role of feedback control in navigating these hybrid systems and placing another steppingstone toward their practical applications.

## 2. Materials and Methods

### 2.1. Animals

*Zophobas morio*, also known as darkling beetles, was used as the living insect platform in this study. The beetle's relatively small size (~2−2.5 cm) and light weight (~0.4−0.6 g) make it a potential mobile platform for developing insect-machine hybrid systems. Although ethical regulations for invertebrate research are still under debate (Freelance 2019), this study attempted to provide good living conditions for these insects. Their colonies were reared inside the compartments of a Mouse Housing System (NexGen® Mouse 500, Allen Town®), allowing clean air to be circulated. The density of each compartment was kept at 20 beetles/~9400 $cm^3$ (i.e., 19 × 13 × 38 $cm^3$), providing a spacious territory. The compartments were washed weekly to maintain their hygiene. Temperature and relative humidity were maintained at ~25 °C and 60%, respectively. Water and food were supplied to the territories twice per week via various vegetables (e.g., carrots and apples). These conditions were provided not only for the intact insects (those to be used for the study) but also for post-experiment ones (those no longer used for the study).

### 2.2. Wireless Backpack Stimulator

A miniature stimulator, or backpack (15 × 5 $mm^2$), was designed to wirelessly control the beetle's locomotion (Fig. 1C, Supplementary Video S1). The backpack utilized a microcontroller CC2650 as its main core (TI, 48 MHz, 128 KB of Flash, Bluetooth® Low Energy 4.2). The backpack allowed a transmission distance of 10 meters and provided up to eight independent stimulation terminals. It was powered by a rechargeable lithium-ion battery (1.5 V, 8 mAh). The total weight of the backpack and battery was ~0.5 g. The backpack was mounted onto the beetle's elytra using beeswax (Figs. 1A, 1B). Its communication with the main PC was mediated by a Bluetooth central station (TI, SmartRF06 Evaluation Board, Supplementary Fig. S1). The backpack was implemented to demonstrate the wireless locomotion control of the beetle via its antennae stimulation (Supplementary Video S1).

*2.3. Implantation and Electrical Stimulation*

Five electrodes made of coated copper wires (55 μm bare diameter, 44 AWG, Remington Industries) were implanted into the beetle's antennae and elytra to induce turning and accelerating motions, respectively (Fig. 1B). The insulation layer at two ends of each electrode was first removed using flame (the de-insulated length was ~1 mm). One of these ends was inserted into the beetle, while the other was connected to an electrical stimulator. Prior to the implantation, the beetle was anesthetized using $CO_2$. The flagellums of its antennae were then trimmed off to insert two working electrodes. The insertion ceased once reaching the antennal base, indicated by a small resistance. A beetle pin (No. 00, Indigo Instruments) was then used to pierce three tiny holes on the beetle's body, one was at the pronotum, and two were at the leading edge's vein of each elytron. Three remaining electrodes were implanted into these opens, in which the common electrode was situated at the pronotum. The depth of these implants was at ~2 mm. All five electrodes were secured using beeswax. The beetle was then given ~1 hour for recovery.

The elytra stimulation was conducted using a fixed electrical stimulus, which had 2.5 V amplitude, 20 Hz frequency, 200 ms duration, and 50% duty cycle. These parameters were selected to attain significant responses from the beetle. The antennae stimulation was performed using a short electrical stimulus, whose amplitude, pulse width, and duration were fixed as 2.5 V, 2 ms, and 400 ms, respectively. These parameters were observed as providing the beetle with apparent turning reactions. The stimulation frequency was adjusted by the navigation program (Fig. 2A). The adjustment range was from 10 Hz to 40 Hz so that the induced angular speed was proportionally graded (Fig. 1E, Supplementary Fig. S2, Supplementary Video S1) (Vo Doan et al. 2017).

*2.4. Setup and Evaluation of Navigation Experiment*

The automatic navigation of the beetle was evaluated under a loosened tethered condition to maintain the experiments for long periods. A long copper wire (i.e., 1.5 meters, 44 AWG Magnet Wire, Remington Industries 44SNSP) was used to transmit the electrical stimuli generated by the stimulator to the beetle. The stimulator was an Arduino Uno board (ATmega238, 16 MHz, 32 kB of Flash) associated with a resistor voltage divider to regulate the stimulus amplitude to 2.5V. This circuit was connected to the main PC via serial communication (460800 bps). The navigation program was written in MATLAB® and embedded into the main PC. The program computed the stimulation commands and transferred them to the stimulator to generate corresponding electrical pulse trains, thus maneuvering the beetle's motion. This computation was processed under a feedback control manner, meaning the program established the command based on the beetle's instantaneous position (Fig. 2A, Supplementary Fig. S1). The position and orientation of the beetle were provided by a 3D motion capture system (Vicon®, 100 fps, 120 × 120 × 120 $cm^3$, Supplementary Fig. S1A), which tracked a set of three retroreflective hemisphere markers affixing on a light carbon fiber frame mounted on the beetle (3 mm radius, ~20 mg, Supplementary Fig. S1B). The coordinates of the three markers were used as the program's feedback data (Supplementary Fig. S1C). They were also synchronized with the stimulation commands and logged by the main PC for post-experimental analysis.

In each navigation trial, the navigation program automatically maneuvered the beetle to move from the origin to the destination, two circular areas with a radius of 40 mm (Fig. 2C). The beetle's journey was expected to follow a predetermined sine curve, whose equation was $y = 170(2\pi x/850)$ (mm) (Fig. 2C).

A trial would be terminated if the beetle reached the destination or left the floor (120 × 60 $cm^2$) or the experimental time exceeded 5 minutes, whichever came first. The destination and origin were swapped between two consecutive trials to avoid potential biases. There were 12 trials conducted for each beetle. An interval of 5 minutes of rest was applied after each trial. A trial was counted as successful if the beetle arrived at the destination before the termination conditions were met. Otherwise, it was recorded as a failed navigation.

The operation of the established feedback system was governed by two control parameters, the proportional gain ($K_p$) and the update interval ($t_{update}$) (Fig. 2A). Its navigation performance was thus evaluated under different combinations of these two parameters. The proportional gain was set to either 0.25, 0.50, or 0.75, whereas the update interval was selected at 1.0 s, 1.5 s, or 2.0 s. The nine combinations were randomized throughout the experiment to avoid biases. Herein, the adjustment of the two control parameters did not serve to find their optimal values but to evaluate the navigation across various conditions, e.g., from frequently monitoring the beetle (fast $t_{update}$) to maximizing its free motion (slow $t_{update}$).

*2.5. Data Analysis*

The navigation performance of the established feedback system was evaluated via the four factors, "success rate," "tracking error," "navigation time," and "control effort" (Figs. 3B to 3E). One-way ANOVA test (0.05 significant level) was employed to study the two control parameters' effect on these factors, i.e., the impact of adjusting $K_p$ under a given $t_{update}$. In addition, statistical comparisons in the study were examined via t-test (0.05 significant level).

The beetle was observed to become unresponsive when several consecutive stimuli were applied to only one of its antennae, which might be a consequence of habituation, implant impairments (caused by accumulated electrical charges), or a contribution of both (Latif and Bozkurt 2015; Rankin et al. 2009). This bias was removed from the data analysis by excluding those trials containing 15 or more consecutive unilateral stimuli. These trials occupied 30/228 total experimented trials, i.e., ~13%. In addition, 41 other trials (~18%) were also excluded because the three retroreflective markers were miss-tracked more than 20% of the time.

Location data of the beetle was passed through a moving average filter (0.1-seconds sliding window) for noise removal. The data was then implemented to calculate the four navigation factors and reconstruct the beetle's motion, i.e., its linear and angular speeds induced by the stimulation (Figs. 1D to 1F, Supplementary Fig. S2), its instantaneous linear speed and distance to the path during the navigation (Supplementary Figs. S4B, S4C). The last two elements were sampled at a rate of 500 ms. The instantaneous linear speed was defined as the average speed within a 100-ms window. The right (i.e., clockwise) turns were assigned negative values, and vice versa.

The beetle's turning response to its antennae stimulation (Figs. 1D, 1E, Supplementary Fig. S2) was reconstructed from the navigations attained with $K_p = 0.50$, which contained diverse stimulation frequencies. The data were then grouped into four ascending groups of the stimulation frequencies, including 10-16 Hz, 17-24 Hz, 25-32 Hz, and 33-40 Hz. Then, outliers were removed and determined as those data falling outside the range of ±2.7σ, where σ was the standard deviation of each group.

In this study, figures, data, and data variability were analyzed and displayed as mean ± standard deviation.

## 3. Results and Discussion

*3.1. Locomotion Control of The Terrestrial Beetle*

Walking control of the terrestrial cyborg beetle was attained via electrical stimulation of its sensory organs (Latif and Bozkurt 2012; Nguyen et al. 2020; Vo Doan et al. 2017). Particularly, the stimulation of an antenna steered the beetle contralaterally, i.e., electrically stimulating the right antenna induced left turns, and vice versa (Figs. 1B, 1D) (Latif and Bozkurt 2012; Vo Doan et al. 2017). This locomotory reaction resembles the beetle's natural response when one of its antennae is suddenly disturbed, e.g., being abruptly touched (Ye et al. 2003). In specific, such a sudden disturbance excites the mechanoreceptors distributed around the antennal appendages, which signal the beetle's CNS (Comer and Baba 2011; Comer et al. 2003). The CNS could then interpret the disturbance as an alert of potential dangers or obstructions and drive the beetle toward the opposite direction to escape the threat or avoid the obstacles.

In addition, the beetle's turning motion was graded by adjusting the frequency of the electrical stimulation (Fig. 1E). Higher frequencies induced faster turning speeds (Supplementary Fig. S2) (Vo Doan et al. 2017). As a result, the induced turning angles became more prominent as the stimulation frequencies were raised (Fig. 1E,

Supplementary Video S1). For example, increasing the stimulation frequencies from 10-16 Hz to 33-40 Hz doubled the elicited turning angle from ~12.5 deg to more than ~25 deg (Supplementary Table S1). Such a graded locomotory response was implemented in the feedback control-based navigation to accurately steer the terrestrial cyborg beetle (Fig. 2A).

In addition to the steerability (i.e., angular control), the electrical stimulation could provide the control of translational motions for the cyborg beetle, e.g., forward/backward movements, left/right sideways walks (Nguyen et al. 2020; Vo Doan et al. 2017). In specific, the elytra stimulation of the beetle accelerated it forward (Nguyen et al. 2020) (Figs. 1B, 1F). Once the electrical signal was transferred to both elytra simultaneously, the beetle rapidly accelerated forward with its linear speed increasing ~40 mm/s averagely (Fig. 1F). Thus, such response was exploited as a thrust controller by the feedback control system to quickly direct the beetle toward frontal destinations (Fig. 2A). Although the graded control of this motion was reported to be inversely proportional to the stimulation frequency (Nguyen et al. 2020), it was not implemented in the thrust controller for the sake of simplicity (Fig. 2A).

*3.2. Feedback Control-Based Automatic Navigation of The Cyborg Beetle*

The feedback control system was designed to automatically navigate the cyborg beetle along a defined path (Figs. 2A, 2C). The system first computed the orientation error between the beetle and its assigned target (i.e., the angle $\theta$, Fig. 2B). This computation was updated periodically at every $t_{update}$ (s). Afterward, a navigation command (i.e., left/right turn or acceleration) was established. The beetle was steered toward the destination when the angle $\theta$ was significant (i.e., $\theta > 25$ deg (Whitmire et al. 2013)). Otherwise, it was accelerated forward. Due to the graded response in turning (Fig. 1E), the antennal stimulation frequency was regulated via a proportional controller, expecting to accurately and quickly reorient the cyborg beetle (Fig. 2A). The controller's proportional gain ($K_p$) and the update interval ($t_{update}$) were altered to evaluate the system's performance. The automatic navigation was executed along a predetermined sine curve (Fig. 2C). This path-following process was accomplished via a carrot-chasing fashion that the beetle was directed toward sequentially generated temporal targets (Fig. 2B). Once the beetle nearly arrived at a target, a novel one approaching the end of the curve was established. This process repeated until the beetle reached its ultimate destination.

Besides this artificially designed system, the beetle intrinsically possesses a complex and naturally evolved feedback control system (Fig. 2A). During its locomotion, the beetle constantly acquires not only internal feedback from proprioceptors (for precise motor control) but also external information via a rich collection of sensory organs distributed over its body, e.g., tactile feedback from its legs, antennae, and elytra, as well as visual input from its eyes (Dickinson et al. 2000; Ritzmann and Büschges 2007). The collected information interacting with the internal locomotory neural circuits (e.g., central rhythm generators) allows the beetle to adjust its movements accurately (Bidaye et al. 2018) and thus adapt to the dynamic surrounding terrains. Cockroaches, for example, reportedly used antennal and visual feedback to detect and negotiate with frontal obstructions *(Blaberus discoidalis)* (Harley et al. 2009) or adjusted their walking gaits according to the slipperiness of surfaces underneath (*Nauphoeta cinerea*) (Weihmann et al. 2017). This intrinsic feedback control system potentially provides the cyborg beetle high robustness and adaptivity, allowing its automatic navigation to be vigorous under external disturbances (e.g., obstacles (Tran-Ngoc et al. 2021)). Thus, combining this inner natural control system with the external artificial controllers would enable robust navigation with high precision and reliability in cyborg beetles. In other words, these two systems could work in tandem and complement each other (Fig. 2A) and thus express the beneficial fusion of engineering and biological perspectives in (terrestrial) cyborg beetles (Tran-Ngoc et al. 2021).

The cyborg beetle was navigated successfully to follow the predetermined sine curve and arrive at the destination (Figs. 2C, 2D, Supplementary Fig. S3, Supplementary Videos S2, S3). In specific, such navigations occurred for ~71% of the time (N = 19 beetles, n = 112/157 trials), combining all sets of $K_p$ and $t_{update}$. Furthermore, the successful navigation trajectories were distributed spatially in the forms of sine curves around the predetermined one (Fig. 2D), fairly reflecting the accuracy and efficiency of this feedback control system. The navigation of contemporary terrestrial insect-machine hybrid systems was relatively challenging, with the insects being successfully navigated along predefined paths for only ~10-30% of the time (Latif and Bozkurt 2012; Whitmire et al. 2013). Despite being just a relative comparison (as navigation was not in these studies' interest), the success rate difference provides a glimpse at a beneficial attribute of the designed system, i.e., the regulation of electrical stimulation. As introduced, the previous works implemented an identical stimulus for different insect individuals despite their physiological distinctions (Erickson et al. 2015; Latif and Bozkurt 2012; Whitmire et al. 2013). This open-loop control greatly diverged the insects' reactions, leading to a significant variation in the controllability and thus lessening the navigation outcomes. Herein, as such a variation was compensated via the feedback regulation of the stimulation frequency (Fig. 2C, Supplementary Video S3), the successful automatic navigation of the cyborg beetle was more likely to be attained (i.e., ~71%, Fig. 2D, Supplementary Fig. S3).

*3.3. Evaluation and Optimization of the Automatic Navigation*

Besides its general success discussed above, tuning the two control parameters (i.e., $K_p$ and $t_{update}$) altered the performance of the feedback system (Fig. 3). For instance, with $t_{update} = 1.0$ s, the beetle stayed closer to the defined path when $K_p$ was set as 0.5 (Fig. 3A). More specifically, adjusting these two parameters impacted four different aspects of the system, including "success rate," "tracking error," "navigation time," and "control effort" (Figs. 3B to 3E). As an indicator of its reliability, the system's success rate was judged as the fraction of successful navigations (Fig. 3B). Similarly, the system's accuracy was expressed via its tracking error. This factor was calculated by dividing the region formed by the beetle's actual trajectory and the predetermined sine curve by the latter's length (Whitmire et al. 2013) (Figs. 2C, 3C). The third factor, navigation time, depicted the time efficiency of the system and was computed as the elapsed time the beetle needed to reach the destination (Fig. 3D). Finally, the system's cost or its control effort was the average number of stimuli it delivered to obtain a successful navigation trial (Fig. 3E). The variations of these four aspects corresponding to the two control parameters would provide the cyborg beetle's automatic navigation the flexibility to be tuned or optimized depending on requirements.

When highly successful navigation is favored over the other three aspects, either a frequent update interval or a large proportional gain should be set (Fig. 3B, Supplementary Table S2). For example, when the beetle was under frequent control (i.e., $t_{update} = 1.0$ s), a high success rate of more than 75% was attained regardless of $K_p$ (Fig. 3B). Especially, a success rate of up to ~94% was recorded using $K_p = 0.5$ under this condition (Fig. 3B). However, when the update interval was increased (i.e., 1.5 s and 2.0 s), leading to a growth in the beetle's control-free motion, the success rate was more significant with the large $K_p$ of 0.75 compared to the other two values (i.e., 0.25 and 0.5) (Fig. 3B). At $t_{update} = 1.5$ s, this $K_p$ resulted in a 20% higher success rate than its two counterparts. The distribution of stimulation frequencies generated under each $K_p$ accounted for these success rate differences. As the proportional gain was large (i.e., $K_p = 0.75$), the frequency was biased toward its high values (Fig. 3F). This tendency steered the beetle with higher angular displacements (Fig. 1E, Supplementary Fig. S4A) and thus compensated the control-free motion.

Unlike the success rate, when $t_{update}$ was 1.0 s, the tracking error was found to be significantly affected by altering $K_p$ (Figs. 3A, 3C, ANOVA test, $P = 0.032$, $df = 45$, Supplementary Table S2). Setting $K_p$ to 0.50 provided an error of 4.99 ± 3.90 mm, whereas that of the other two gains was ~4 mm higher. A different aspect of the navigation accuracy was the beetle's distance to the sine route during its journey (Fig. S4B, Supplementary Table S2). When $K_p$ was set as 0.50, this distance was 9.82 ± 6.34 mm. It, however, became further when the other two $K_p$ was used (Fig. S4B, t-test, $P < 0.001$, $df > 2000$). The superiority of $K_p = 0.50$ might relate to the diversity of its stimulation frequencies (Fig. 3F), which almost evenly spread from 10 Hz to 40 Hz. Meanwhile, the frequencies employed by the other two gains were slanted toward either end of the spectrum. The lean toward low frequencies (i.e., $K_p = 0.25$) might result in insignificant effects on correcting the beetle's orientation, whereas the other extreme (i.e., $K_p = 0.75$) possibly overshot the beetle, fluctuated it along the path, and thus worsen the navigation (Fig. 3A, Fig. S4A). Unlike the success rate, the impact of $K_p$ on the tracking error no longer presented when the beetle's control-free motion became dominant (i.e., $t_{update}$ = 1.5 s and 2.0 s, Fig. 3C, ANOVA test, $P > 0.1$, $df > 30$, Supplementary Table S2).

Despite its high success rate and low tracking error, the combination of $K_p$ = 0.50 and $t_{update}$ = 1.0 s did not benefit the navigation program in terms of control effort. Averagely, the program applied ~37 stimuli to successfully direct the beetle using this combination, which was not an advance compared to the other two gains (Fig. 3E, t-test, $P > 0.6$, $df > 27$, Supplementary Table S2). As all three $K_p$ used the same update interval of 1.0 s, their similarity in control effort was consistent with their insignificant effect on the navigation time (Fig. 3D, ANOVA test, $P > 0.1$, $df > 30$, Supplementary Table S2). However, when $t_{update}$ was raised, large proportional gains such as 0.50 and 0.75 tended to favor these two navigation factors (Figs. 3D, 3E, t-test, $P < 0.04$, $df > 18$). In the instance of $t_{update}$ = 2.0 s, the necessary number of stimuli for a successful trial increased from ~30 to ~51 when $K_p$ was reduced from 0.75 to 0.25. This tendency was possibly associated with the stimulation frequency, in which the beetle was more energized under high frequencies (established by $K_p$ = 0.50 and 0.75), leading it to reach the destination earlier. Although these stimulations were meant to steer the beetle, this argument was not invalid. The beetle's reactions are not pure rotations but accompanying forward runs, which is a common escape strategy of terrestrial insects (Domenici et al. 2008). In addition, a closer look at the beetle's instantaneous linear speed throughout its navigation also supported the argument (Supplementary Fig. S4C, Supplementary Table S2). At $t_{update}$ = 1.5 s, for example, the speed was ~11 mm/s with $K_p$ = 0.25, and then increased to ~24 mm/s and ~16 mm/s with $K_p$ = 0.5 and 0.75, respectively (Fig. S4C, t-test, $P < 0.001$, $df > 2000$).

*3.4. Essential Role of Feedback Control for Terrestrial Cyborg Insects*

Herein, the experimental outcome indicates improved reliability in the navigation of terrestrial cyborg insects. The success rate in path-following control is increased from ~30% to more than 90%, loosely compared to existing studies (Latif and Bozkurt 2012; Whitmire et al. 2013). Such an improvement was achieved with a shift in control philosophy. Instead of considering the cyborgs as purely biological subjects, the proposed navigation system views them from an engineering perspective and applies a feedback control-based approach to their navigation. As introduced, existing navigations of these cyborgs tend to lack reliability due to the high variation in their locomotory responses and the maneuvering difficulties under manual controls. While the latter can be eliminated by introducing automatic navigations, the former remains a challenge. As each insect is biologically unique, it is not easy to establish an optimal stimulus that minimizes the high variation in responses between individual insects (Erickson et al. 2015). Herein, instead of seeking such an optimal signal, the proposed system strategically adjusts the stimulation according to the insects' instantaneous locomotory status. This feedback control-based strategy is shown to be efficient, with highly successful and accurate navigations being attained. Furthermore, such a navigation efficiency is also in line with other successful utilizations of feedback systems in controlling cyborg insects, e.g., inflight control and leg motion regulation in giant flower beetles (*Mecynorrhina torquata*) or orientation adjustment in migratory locust (*Locusta migratoria manilensis*) (Cao et al. 2014; Li et al. 2018; Liu et al. 2022). Together, these results consolidate the essential role of feedback systems, or engineering control perspectives, for the reliable control/navigation of terrestrial cyborg insects.

Besides solidifying the role of feedback control-based approaches, the experimental outcome also provides insight into studies on the electrical stimulation of living insects. In specific, as reliable and accurate navigations could be attained without an optimal stimulus, the quest to search for such stimulation should be adjusted. Instead of exclusively concentrating on optimizing the stimulation (Erickson et al. 2015), upcoming studies might complement such a task by investigating possible relationships between the electrical stimulation and the insects' induced locomotory responses. Those relationships will be a significant foundation for developing applicable feedback control systems, which tactically alter the stimulation parameters rather than fixing them to return more efficiently navigating/controlling outcomes. For example, the proportional controller presented herein is based on the graded turning response of darkling beetles to their antennae stimulation (Vo Doan et al.

2017). A similar graded reaction can be popularly found in various insect species. E.g., it appears in the electrical stimulation of Madagascar hissing cockroaches' sensory organs (*Gromphadorhina portentosa*) or giant flower beetles' muscular systems (*Mecynorrhina torquata*) (Cao et al. 2014; Latif et al. 2016; Li et al. 2018). Thus, it is practically ambitious to achieve precise and highly successful navigations across the diversity of the insect kingdom by implementing feedback control-based approaches.

Together with reliability and precision, the experimental outcome implies that the navigation of terrestrial cyborg insects can possess another essential feature owing to applying the feedback control-based approach, the tunable performance. As shown, the insects' navigation depends on not only their physiology but also the implemented controller and its parameters. The parameters can be adjusted to serve different purposes. For example, accurate path-following navigations would need a fast $t_{update}$ and a medium $K_p$, whereas navigations expecting only a high success rate would be attained with a large $K_p$ disregarding $t_{update}$. Such performance tunability will help to provide flexibility to the cyborg insects' practical applications. For instance, a cyborg insect operating in an extended exploration mission might dial up its $K_p$ to maintain reliable navigation while reducing its position update rate (i.e., $t_{update}$) for power-saving purposes. Like their parameters, different controllers will satisfy different goals. For instance, the proportional controller proposed herein is deployed to favor the tracking error, making it mostly irrelevant to the navigation time. Thus, an additional proportional controller for the forward acceleration (i.e., the elytra stimulation) might be necessary for compensation. Besides, more advanced controllers, e.g., PID controller, Fuzzy Logic, or Deep Learning, can be considered to reduce errors and improve navigation performance (Liu et al. 2022; Yang et al. 2022; Zhang et al. 2016). As aforementioned, realizing terrestrial cyborg insects' practical uses requires more effort, and feedback control-based approaches can be a helpful tool for this endeavor.

### 3.5. Limitations and Future Works

Expanding the proposed feedback system with more advanced control techniques, e.g., a PID controller, might benefit navigation performance. However, such an expansion should consider the cost-and-benefit trade-offs. For example, adding the derivative and integral terms might quickly increase the cost and time consumption to fine-tune the system's control parameters. Unlike regular robotic systems, cyborg insects are not yet equipped with efficient tools for PID tuning (Borase et al. 2021; Tan et al. 2006). Existing applications of the PD controller for these cyborgs still limit their analysis to only one set of empirically predetermined derivative and proportional gains (Li et al. 2021; Li et al. 2018; Liu et al. 2022). Herein, the reported parameters for highly successful and accurate navigations (i.e., $K_p$ of 0.5 and $t_{update}$ of 1.0 s) are also tuned heuristically instead of based on more efficient methods, e.g., the model-based tuning approach (Iplikci 2010). Thus, to have a robust foundation to effectively study the use of advanced control techniques in cyborg insects, the next logical step might be building behavior and control models of the insects based on the relationship between the electrical stimulation inputted and the insects' locomotory responses outputted (Holzer and Shimoyama 1997). In addition, as this relationship most likely reflects a nonlinear and time-variant process (Zhang et al. 2016), nonlinear control theories, e.g., Fuzzy Logic (Zhang et al. 2016), should be considered together with those linear ones (e.g., PID controller). Similarly, the use of AI, e.g., Deep Learning (Yang et al. 2022), in regulating the cyborgs' locomotion should not be overlooked if the computational resource can be well managed.

Although the engineering control perspectives benefit navigation performance, the actualization of terrestrial cyborg insects' practical uses can't be completed if disregarding other vital aspects of the insects, especially their biological and physiological characteristics. Herein, the experimental outcome experiences a gradual attenuation of the darkling beetle's reaction to the electrical stimulation (Supplementary Fig. S5). That means the feedback system tends to close its control of the cyborg beetle over time, making later navigations more easily prone to failure than those conducted earlier (Supplementary Fig. S5). Such a response attenuation might result from habituation, impairments of the tissue-electrode interface, or a combination of both (Latif and Bozkurt 2015; Rankin et al. 2009). In general, these issues can't be solved by relying on only engineering perspectives but request closer looks at the insects' biological and physiological aspects. For example, to minimize the tissue damage due to implanted electrodes and enhance the interface's bioelectrical properties, the implantation should be carried out during the insects' metamorphic stage (Bozkurt et al. 2011). In addition, using inert and biocompatible materials, e.g., platinum, as the electrodes can help reduce toxic chemical reactions occurring during the stimulation, thus preventing the interface's injuries (Merrill et al. 2005). Furthermore, the impedance and water window of the interface can be measured as indicators for excessive stimulation, which potentially results in its impairments (Latif and Bozkurt 2012; Latif and Bozkurt 2015). Besides, the insects might less habituate and respond more naturally to certain stimulation waveforms, e.g., those in the forms of multiple bursts (Jamali et al. 2019), suggesting the necessity of studies on insect-friendly stimulation protocols. Due to their biohybrid nature, the development of terrestrial cyborg insects will request studies on both of their facets, i.e., the engineering and biological characteristics.

## 4. Conclusion

This study evaluates and solidifies the role of feedback control in the automatic navigation of terrestrial insect-machine hybrid systems. This approach helps overcome the issue of high variation in responses of individual insects, which long obstructed the navigation control of ambulatory insect-machine hybrid systems. In addition, the feedback control-based navigation is shown to be precise, reliable, as well as flexibly adjustable. Although there are still many technical obstacles to overcome to bring the ideas of insect-machine hybrid systems from laboratory environments to practical applications, e.g., the need for an onboard localization system and well-planned power management, the results of this study contribute an essential step toward this challenging but intriguing journey.

**CRediT authorship contribution statement**

**H. Duoc Nguyen**: Conceptualization, Software, Methodology, Investigation, Writing - Original Draft. **V. Than Dung**: Software. **Hirotaka Sato**: Resources, Funding acquisition, Supervision, Project administration. **T. Thang Vo-Doan**: Conceptualization, Methodology, Supervision, Project administration, Writing - Original Draft.

**Declaration of competing interest**

The authors declare that they have no known competing financial interests or personal relationships that could have appeared to influence the work reported in this manuscript.

**Acknowledgment**

This work was supported by the Singapore Ministry of Education (RG140/20) (Corresponding authors: Hirotaka Sato, T.T. Vo-Doan). T.T. Vo-Doan was supported by Human Frontier Science Program Cross-disciplinary Fellowship. The authors offer their appreciation to Mr. Chew Hock See, Ms. Kerh Geok Hong, Mr. Tan Kiat Seng, Mr. Roger Tan Kay Chia at the School of MAE, NTU, and Prof. Andrew Straw at the University of Freiburg for their continuous support in setting up and maintaining the research facilities.

## Appendix A. Supplementary data

Supplementary data to this manuscript can be found online.

# Figure Lists

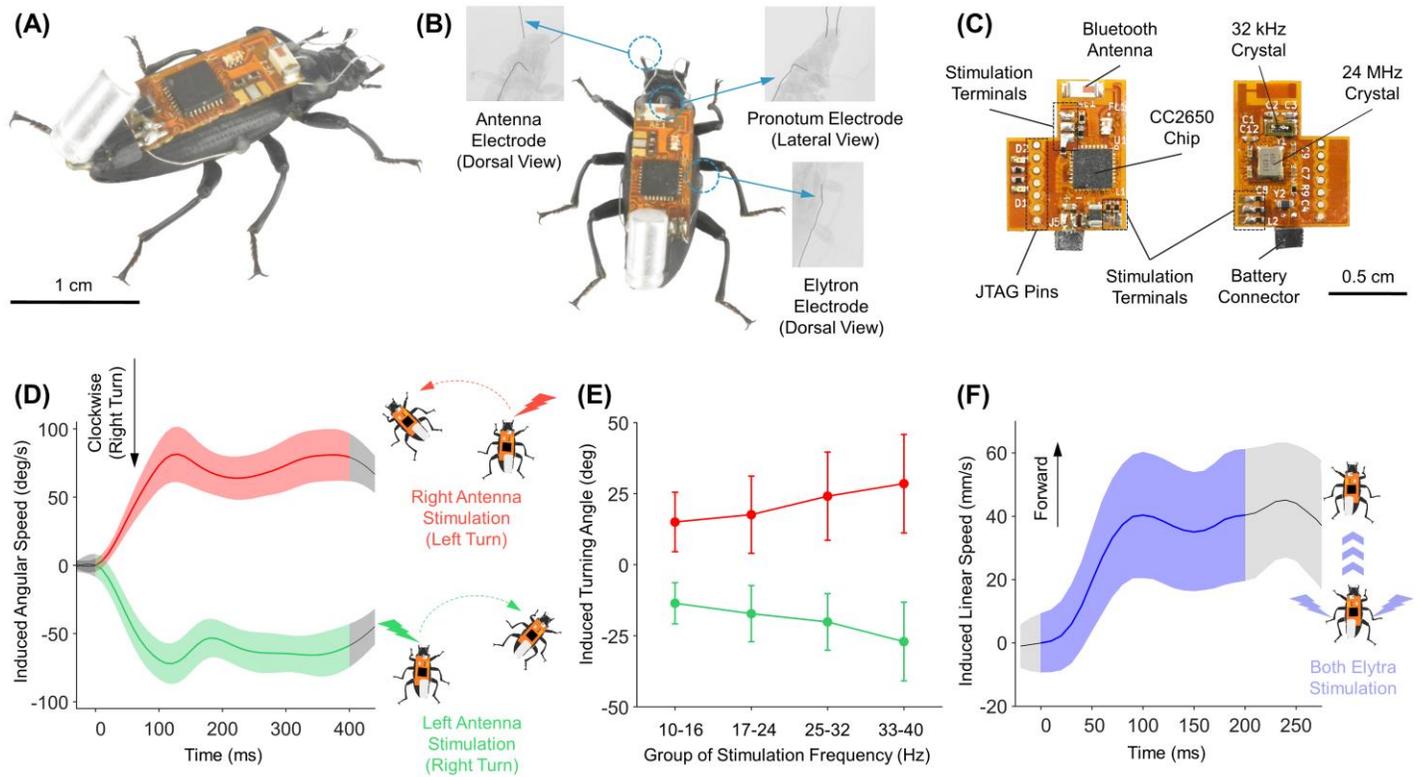

**Fig. 1. Overview of the terrestrial cyborg beetle and its locomotion control.**

(**A**) The cyborg beetle is the fusion of a living ambulatory insect, *Zophobas morio* (~0.6 g), and a wireless stimulator (i.e., the backpack, ~0.25 g) mounted on its elytra. The backpack is powered by a rechargeable Lithium Titanate battery (1.5 V, 8 mAh, ~0.2 g). (**B**) Two pairs of working electrodes are implanted into the beetle's antennae and elytra to induce its turning motions and acceleration. The common electrode is inserted into its pronotum. (**C**) The wireless backpack uses a Bluetooth Low Energy microcontroller as its core and provides up to eight programmable stimulation terminals. (**D**) Representative data of the beetle's turning response. Once the stimulation is applied, the beetle turns contralaterally. Its angular speed rapidly ramps up within the first 100 ms and then fluctuates around a saturated level. Black and colored lines indicate the mean angular speed during stimulation-free and stimulation periods. The shaded regions display standard deviation. (**E**) Graded control of the beetle's turning response via the stimulation frequencies. Higher frequencies tend to elicit faster angular speeds and larger turning angles. Red and green colors represent the right and left antennae stimulation, respectively. Circular markers denote the mean of the induced turning angle, whereas error bars show its standard deviation (N = 16 beetles, n = 859 stimuli). (**F**) The cyborg beetle's forward acceleration is induced by stimulating its two elytra simultaneously. The beetle's forward speed quickly rises when the stimulation is applied. Black and blue lines represent the mean forward speed in stimulation-free and stimulation periods, respectively. The shaded regions show the standard deviation.

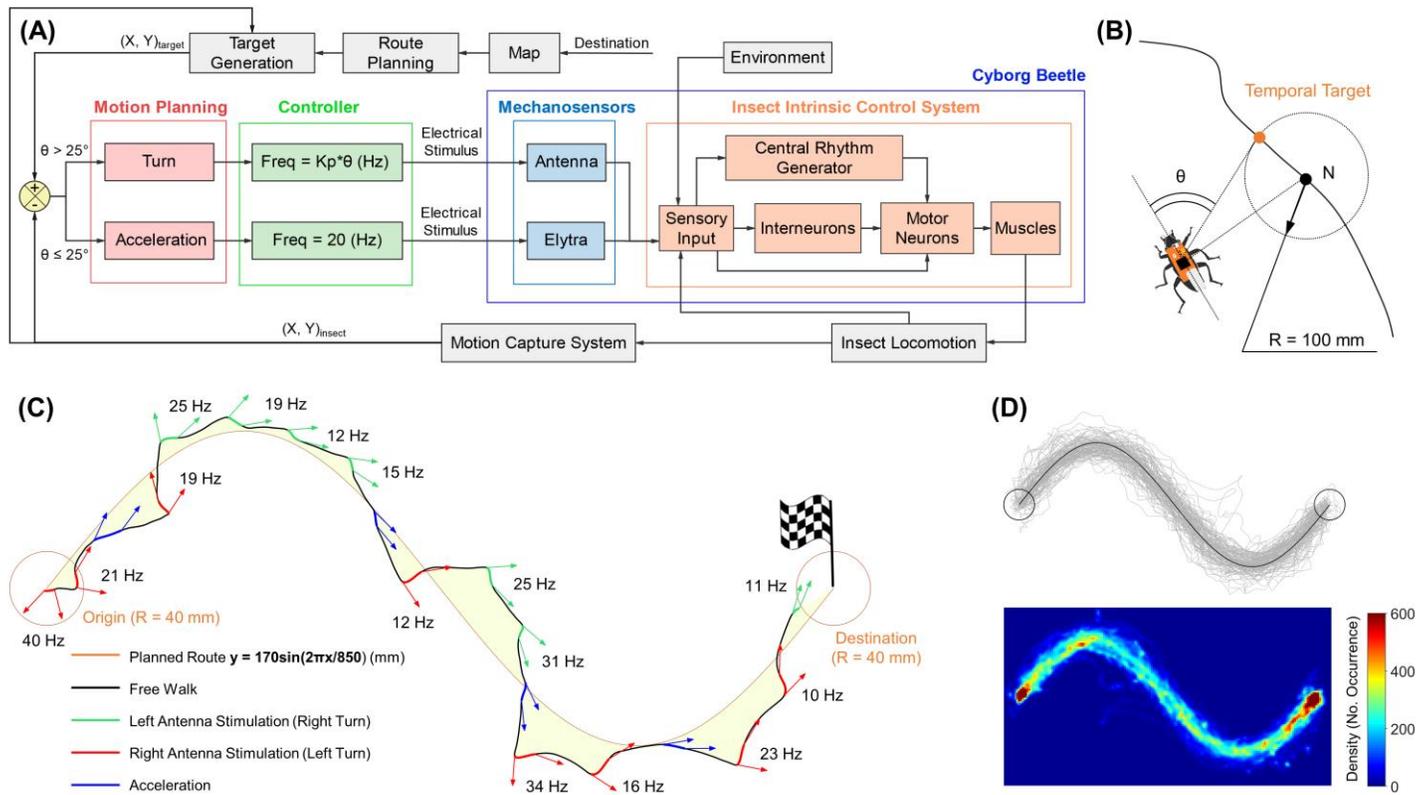

**Fig. 2. The feedback control system for autonomous navigation of the cyborg beetle.**

(**A**) The system combines two artificial control units (i.e., the motion planning block and central controller) and the cyborg beetle's natural control system (i.e., its sensory organs and CNS). First, the motion planning block works out the navigation command (i.e., steering or accelerating) by comparing the beetle's location and its assigned path, which can be planned via the destination and a map of the surrounding terrain. Then, the central controller (i.e., the proportional and thrust controllers) issues an appropriate electrical stimulus to stimulate the beetle's corresponding sensory organs. Finally, the beetle's CNS perceives and uses this electrical stimulus together with other external cues to adaptively alter its motions. The induced locomotion is monitored and feedbacked to the system via a 3D motion capture unit (Supplementary Fig. S1). (**B**) The system attained the path-following control by sequentially generating temporal targets along the path and directing the beetle toward them. These targets are intersections between the circles centering at the beetle's projection onto the path and the path itself (i.e., point N). (**C**) The system attempts to navigate the beetle toward these targets by adjusting the stimulation frequency according to its orientation error, the angle $\theta$. This adjustment prevents the necessity of an optimal stimulus. The shaded yellow region is used to calculate the tracking error. (**D**) The system navigates the beetle successfully ~71% of the time (N = 19 beetles, n = 112/157 trials), marking a significant improvement (loosely compared to previous works). Successful navigation is counted if the beetle reaches the destination within 5 minutes. Gray curves represent the successful trials, whereas the black curve denotes the predetermined path. The heatmap shows the distribution of the beetles' positions in successful navigations.

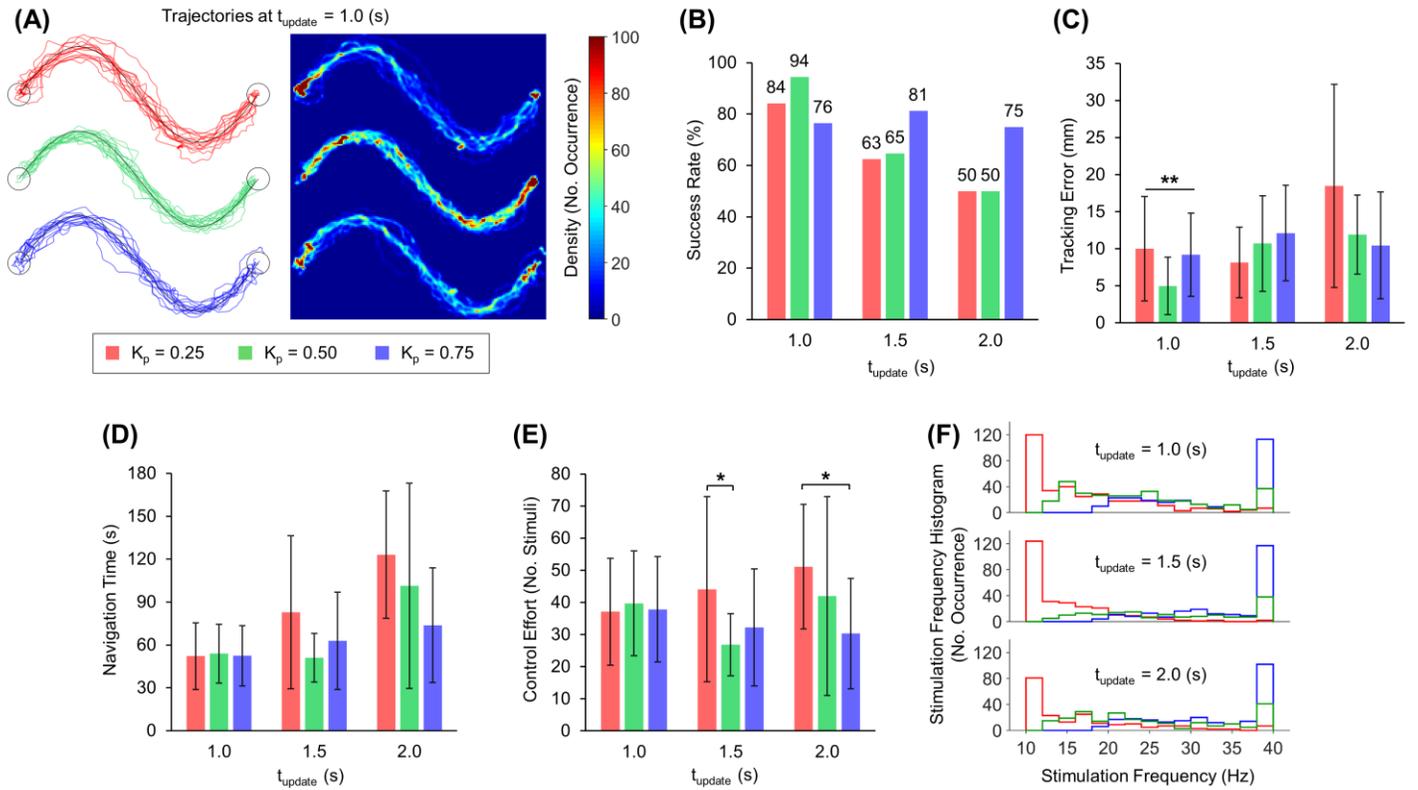

**Fig. 3. Performances of the feedback control system when tuning $t_{update}$ and $K_p$.**

(**A**) The adjustment of $K_p$ and $t_{update}$ enables a flexible regulation of the navigation's reliability (i.e., success rate) and precision (i.e., tracking error). As depicted, the moderate $K_p$ of 0.5 is sufficiently large to correct the beetle's orientation but not too big to cause overshoots. Thus, the beetle's trajectories of this $K_p$ are more spatially concentrated around the predetermined path or more precise than the other two gains. The effect of the two control parameters is apparently displayed via two factors: (**B**) success rate and (**C**) tracking error (\*\*: $P < 0.05$, ANOVA test). The combination of a fast update interval and a moderate $K_p$ allows the cyborg beetle to be navigated precisely (i.e., less than ~10 mm tracking error) and reliably (i.e., more than 90% success rate). However, the two parameters do not significantly affect (**D**) navigation time and (**E**) control effort (although slow $t_{update}$ seems to request large $K_p$ to favor these factors, \*: $P < 0.05$, t-test). This insignificance is expected as the beetle's graded acceleration (Nguyen et al. 2020) is not integrated into the thrust controller. (**F**) Under the proportional controller, the stimulation frequencies steering the beetle are constantly adjusted. The distribution of these frequencies is skewed either toward the left or right end of the frequency spectrum under a minor (i.e., 0.25) or large (i.e., 0.75) $K_p$, respectively. Meanwhile, it is fairly spread when a moderate $K_p$ is selected. These distribution differences help explain the variation in the system's navigation performance. Red, green, and blue represent three values of $K_p$, i.e., 0.25, 0.50, and 0.75, respectively. Error bars display standard deviations. N = 19 beetles, $14 \leq n \leq 20$ trials for each pair of the two parameters.

# Supplementary Figures

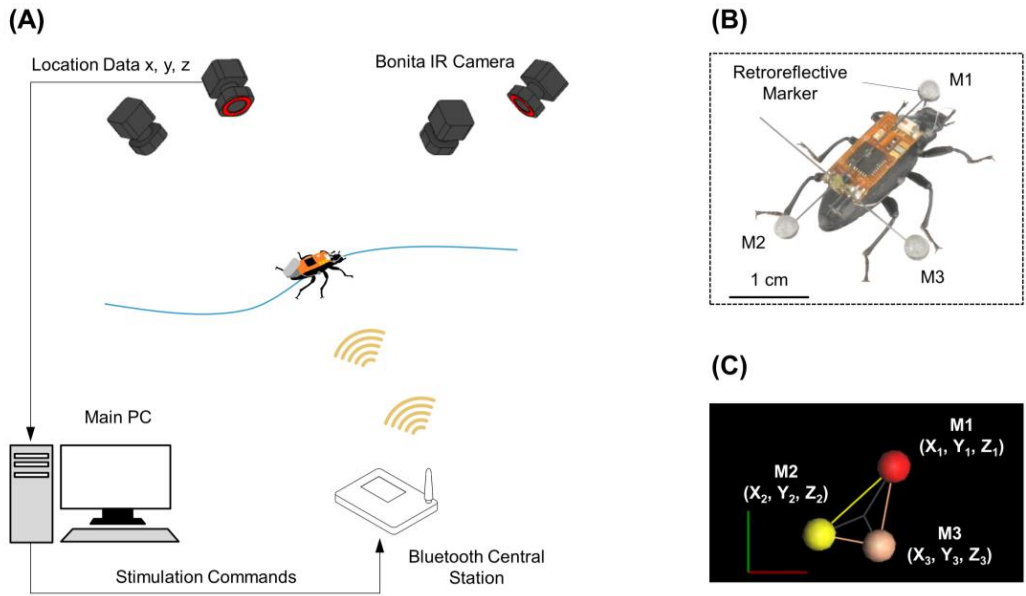

**Fig. S1. Setup of the navigation experiment.**

(**A**) The study is conducted within a 3D motion capture system (Vicon®) made of an aluminum frame and four infrared (IR) cameras (Bonita, 100 fps). The beetle is navigated on a Styrofoam sheet ($120 \times 60$ cm$^2$), following a predetermined sine curve. Each navigation trial is terminated when the beetle is out of the sheet or not reaching the destination after 5 mins, and a failed navigation is recorded. The four cameras monitor the beetle's location and stream it to the main PC to work out the stimulation, which is then sent to the beetle to alter its motion and thus close the control loop. (**B**) Three retroreflective markers are mounted on the beetle's elytra to represent its position and orientation. (**C**) Coordinates of these markers are digitalized using Vicon Tracker® software and transferred to the navigation program embedded in the main PC.

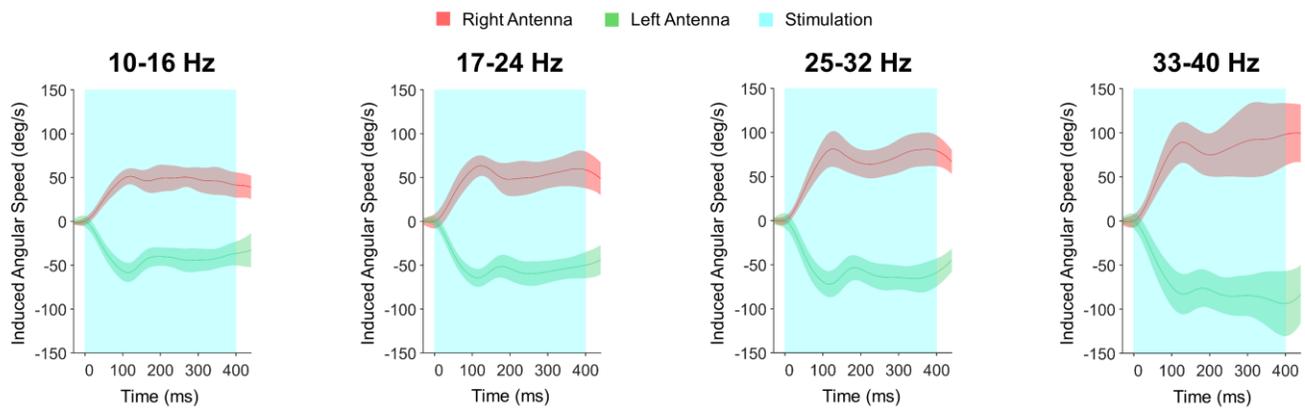

**Fig. S2. Graded response of the beetle to its antennae stimulation.**

The angular speed elicited by the electrical stimulation tends to increase as the stimulation frequency is incrementally adjusted (N = 16 beetles, n = 859 data points). In addition, this speed promptly rises upon the arrival of the electrical stimulus. It then peaks after ~100 ms and slightly fluctuates afterward. As introduced, this graded reaction inspires the implementation of a proportional controller to steer the cyborg beetle.

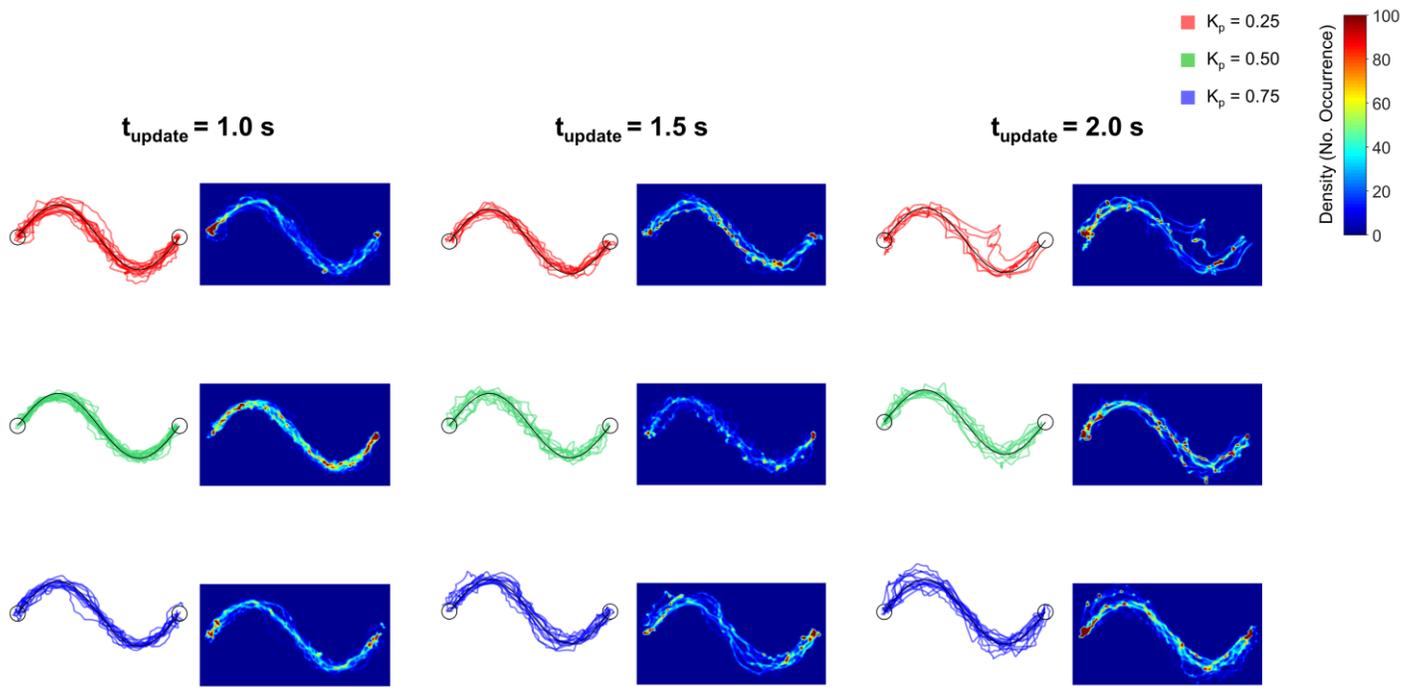

**Fig. S3. Successful navigations in each pair of the two control parameters.**

The adjustment of these parameters impacts the reliability (i.e., success rate) and precision (i.e., tracking error) of the cyborg beetle's navigation, implying the beetle operates similarly to a traditional engineering control system (N = 19 beetles, $14 \leq n \leq 20$ trials for each pair). For example, a fast update interval ($t_{update}$ = 1.0 s) can be paired with a moderate $K_p$ of 0.5 to attain reliable and precise navigations. In contrast, slow update intervals (i.e., 1.5 s or 2 s) need a large $K_p$ to guarantee a high success rate.

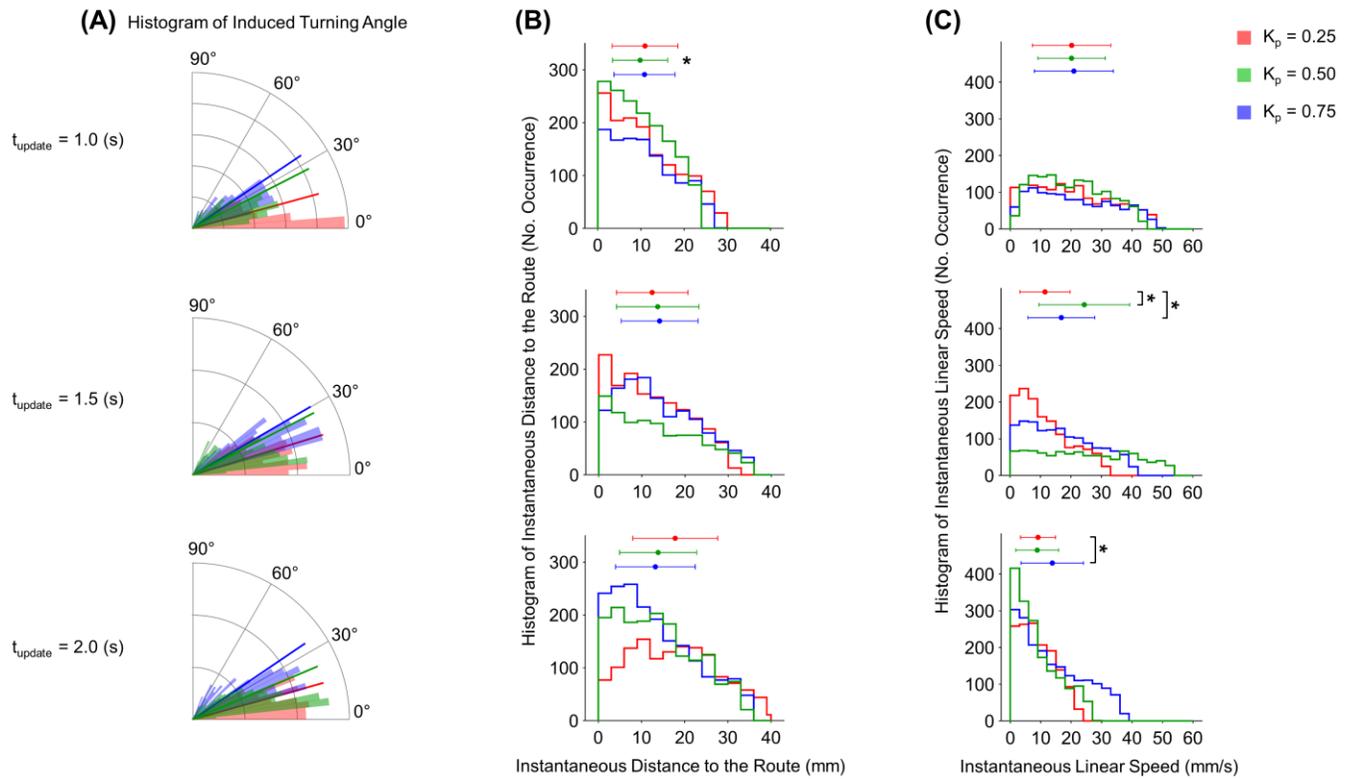

**Fig. S4. Different looks at the feedback-based control navigation.**

(**A**) In tandem with the frequency distribution (Fig. 3F), the mean of induced turning angles in large and small $K_p$ is either larger or smaller than a moderate $K_p$, helping to explain their imprecise navigations given a short update interval ($t_{update}$ = 1.0 s, Fig. 3C). (**B**) Consistent with the tracking error (Fig. 3C) is the distance of the beetle to the predetermined path during its navigation. Under a frequent control ($t_{update}$ = 1.0 s), a moderate $K_p$ of 0.5 keeps the beetle staying near the path (*$P < 0.001$, t-test), resulting in a small tracking error. (**C**) In line with the navigation time and control effort (Figs. 3D, 3E), $K_p$ has no significant effect on the beetle's instantaneous speed when $t_{update}$ is fast. When $t_{update}$ is slow, larger $K_p$ tends to make the beetle move faster (*$P < 0.04$, t-test), reducing navigation time and control effort.

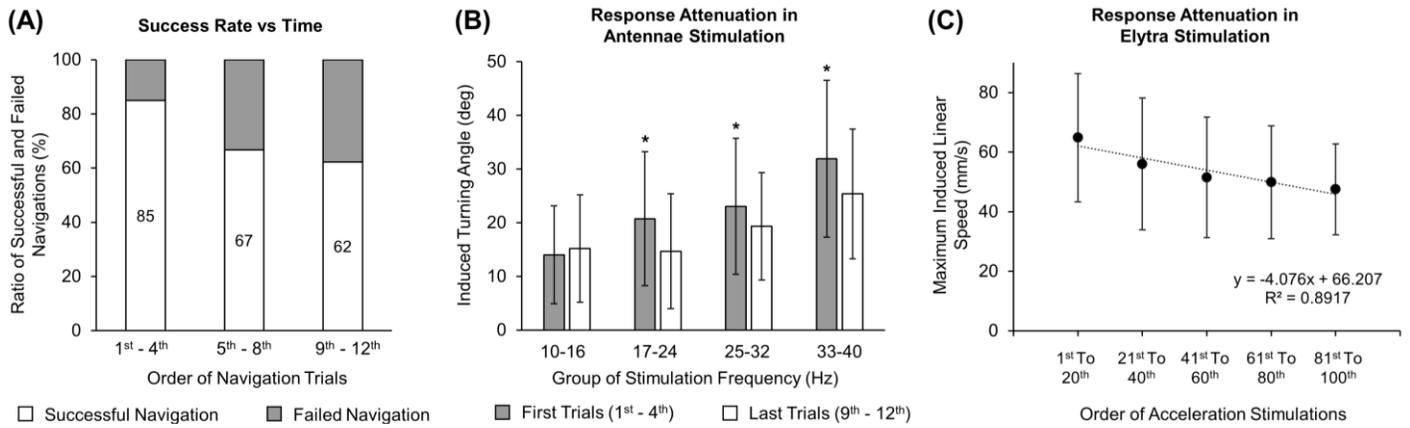

**Fig. S5. Attenuation of the cyborg beetle's response and its impact on navigation performance.**

(**A**) The feedback system is more prone to failed navigations as time passes (N = 19 beetles, 51 ≤ n ≤ 53 trials per column). The likelihood that the cyborg beetle couldn't complete the navigation at the last four (out of twelve) trials increases by nearly 23% compared to those carried out at the first four trials. Such a reduction in navigation reliability results from the attenuation in the beetle's responses to the electrical stimulation. (**B**) In the last four navigation trials, the beetle's elicited turning angle becomes smaller than those induced in the first four trials despite having the same stimulation frequencies (N = 19 beetles, 138 ≤ n ≤ 249 data points per column, *: $P < 0.05$, t-test). (**C**) Similarly, the induced forward linear speed gradually reduces toward the later stimulations. Data is plotted in groups of twenty elytra stimuli and the forms of mean ± standard deviation (52 ≤ n ≤ 64 data points per group). These attenuations cause the feedback system to lose control of the beetle, thus worsening the navigation reliability. Habituation and impairments of the tissue-electrode interface might account for such response attenuations.

## Supplementary Tables

**Table S1. Induced turning angles under the antennae stimulation.** Higher stimulation frequencies likely steer the beetle more profoundly.

| Stimulation Frequency (Hz) | Induced Turning Angle (deg) | |
|---|---|---|
| | Left Antenna (Right Turns) | Right Antenna (Left Turns) |
| 10 - 16 | -13.55 ± 07.25 | 15.01 ± 10.46 |
| 17 - 24 | -17.23 ± 09.88 | 17.60 ± 13.56 |
| 25 - 32 | -20.12 ± 09.95 | 24.11 ± 15.51 |
| 33 - 40 | -27.04 ± 13.84 | 28.50 ± 17.34 |

**Table S2. Variation of the feedback control-based navigation when adjusting the two control parameters.** Owing to the feedback control system, the beetle's navigation outcome can be adjusted depending on applications.

| $t_{update}$ (s) | $K_p$ | Success Rate (%) | Tracking Error (mm) | Navigation Time (s) | Control Effort (No. Stimuli) | Distance to Path (mm/s) | Linear Speed (mm/s) |
|---|---|---|---|---|---|---|---|
| 1.0 | 0.25 | 84 | 10.00 ± 07.04 | 52 ± 23 | 37 ± 17 | 10.95 ± 07.60 | 20.16 ± 12.88 |
|  | 0.50 | 94 | 04.99 ± 03.90 | 54 ± 21 | 40 ± 16 | 09.82 ± 06.34 | 20.12 ± 11.06 |
|  | 0.75 | 76 | 09.22 ± 05.63 | 52 ± 21 | 38 ± 16 | 10.83 ± 07.07 | 20.88 ± 12.96 |
| 1.5 | 0.25 | 63 | 08.17 ± 04.75 | 83 ± 54 | 44 ± 29 | 12.40 ± 08.28 | 11.48 ± 08.26 |
|  | 0.50 | 65 | 10.71 ± 06.45 | 51 ± 17 | 27 ± 10 | 13.67 ± 09.55 | 24.38 ± 14.89 |
|  | 0.75 | 81 | 12.12 ± 06.44 | 63 ± 34 | 32 ± 18 | 14.11 ± 08.92 | 16.84 ± 11.00 |
| 2.0 | 0.25 | 50 | 18.50 ± 13.70 | 123 ± 44 | 51 ± 19 | 17.81 ± 09.85 | 09.11 ± 05.74 |
|  | 0.50 | 50 | 11.91 ± 05.31 | 101 ± 72 | 42 ± 31 | 13.86 ± 08.92 | 08.85 ± 07.04 |
|  | 0.75 | 75 | 10.46 ± 07.22 | 74 ± 40 | 30 ± 17 | 13.22 ± 09.24 | 13.80 ± 10.25 |

**Supplementary Videos**

### Video S1. Graded locomotion control of the insect-machine hybrid system

The graded locomotion control of the beetle is attained by adjusting the stimulation frequency. Higher frequencies tend to induce faster angular speeds. Such a graded response inspires the use of feedback control, e.g., proportional controller, in the navigation of insect-machine hybrid systems.

### Video S2. Feedback control-based automatic navigation of the insect-machine hybrid system

Successful navigations of the beetle are digitally reconstructed from the collected data. For a convenient display, trajectories of these navigations are plotted as if the destination was not swapped between two ends of the predetermined sine path. The variation of the two control parameters provides different performances, i.e., flexibility, to the navigation of insect-machine hybrid systems.

### Video S3. Demonstration of the feedback control-based automatic navigation in a loosely tethered condition

The feedback system alters the stimulation command, i.e., left/right turns or forward acceleration, and its parameter, i.e., the stimulation frequency, according to the locational difference between the beetle and its target on the predetermined path. The system issues a new target once the beetle approaches the old one, resembling a carrot-chasing process, thus attaining path-following control. The demonstration is reconstructed from the collected data with $K_p = 0.25$, and $t_{update} = 1.0$ s.